\documentclass[11pt]{article}
\usepackage{acl2014}
\usepackage{times}
\usepackage{url}
\usepackage{latexsym}
\usepackage{psfrag}
\usepackage{graphicx}
\usepackage{color}
\usepackage[color]{changebar}
\definecolor{light-gray}{gray}{0.33}
\cbcolor{light-gray}

\title{Measuring Word Significance \\ using \\ Distributed
  Representations of Words}

\author{Adriaan M.\ J.\ Schakel \\ NNLP \\ \texttt{adriaan.schakel@gmail.com} \\\And
Benjamin J.\ Wilson \\ Lateral GmbH \\ \texttt{benjamin@lateral.io}} 

\date{today}
\begin{document}
\maketitle

\begin{abstract}
Distributed representations of words as real-valued vectors in a
relatively low-dimensional space aim at extracting syntactic and semantic
features from large text corpora.  A recently introduced neural network,
named word2vec \cite{MCCD13,MikolovSCCD13}, was shown to encode semantic
information in the direction of the word vectors.  In this brief report,
it is proposed to use the length of the vectors, together with the term
frequency, as measure of word significance in a corpus.  Experimental
evidence using a domain-specific corpus of abstracts is presented to
support this proposal.  A useful visualization technique for text
corpora emerges, where words are mapped onto a two-dimensional plane and
automatically ranked by significance.
\end{abstract}

\section{Introduction}
Discovering the underlying topics or discourses in large text corpora is
a challenging task in natural language processing (NLP).  A statistical
approach often starts by determining the frequency of occurrence of
terms across the corpus, and using the term frequency as a criterion for
word significance---a thesis put forward in a seminal paper by Luhn
\cite{Luhn:1958}.  From the list of terms ranked by frequency, terms
that are either too rare or too common are usually dropped, for they are
of little use.  For a domain-specific corpus, the top ranked terms in
the trimmed list often nicely summarize the main topics of the corpus,
as will be illustrated below.

For more detailed corpus analysis, such as discovering the subtopics
covered by the documents in the corpus, the term frequency list by
itself is, however, of limited use.  The main problem is that within a
given frequency range, function words, which primarily have an
organizing function and carry little or no meaning, appear together with
content words, which represent central features of texts and carry the
meaning of the context.  In other words, the rank of a term in the
frequency list is by itself not indicative of meaning \cite{Luhn:1958}.

This problem can be tackled by replacing the corpus-wide term frequency
with a more refined weighting scheme based on document-specific term
frequency \cite{Aizawa:2000}.  In such a scheme, a document is taken as
the context in which a word appears.  Since key words are typically
repeated in a document, they tend to cluster and to be less evenly
distributed across a text corpus than function words of the same
frequency.  The fraction of documents containing a given term can then
be used to distinguish them.  Much more elaborate statistical methods
have been developed to further explore the distribution of terms in
collections of documents, such as topic modeling \cite{Blei:2003} and
spacing statistics \cite{Ortuno:2002}.

An even more refined weighting scheme is obtained by reducing the
context of a word from the document in which it appears to a window of
just a few words.  Such a scheme is suggested by Harris' distributional
hypothesis \cite{harris54} which states ``that it is possible to define
a linguistic structure solely in terms of the `distributions' (=
patterns of co-occurrences) of its elements'', or as Firth famously put
it \cite{firth1957} ``a word is characterized by the company it keeps''.

Word co-occurrence is at the heart of several machine learning
algorithms, including the recently introduced word2vec by Mikolov and
collaborators \cite{MCCD13,MikolovSCCD13}.  Word2vec is a neural network
with a single hidden layer that uses word co-occurrence for learning a
relatively low-dimensional vector representation of each word in a
corpus, a so-called distributed representation \cite{hinton:1986}.  The
dimension is typically chosen of order 100 or 1000.  This is easily
orders of magnitude smaller than the size of a vocabulary, which would
be the dimension when a one-hot representation of words is chosen
instead.  Given the words appearing in a context, the neural network
learns by predicting (the representation of) the word in the middle, or
\textit{vice versa}.  During training, words that appear in similar
contexts are grouped together in the same direction by this unsupervised
learning algorithm.  The distributed representation thus ultimately
captures semantic similarities between words.  This has been
impressively demonstrated by a series of experiments in the original
word2vec papers, where semantic similarity was measured by the dot
product between \textit{normalized} vectors.

In this brief report, we consider the problem of identifying significant
terms that give information about content in text corpora made up of
short texts, such as abstracts of scientific papers, or news summaries.
It is proposed to use the L$_2$ norm, or length of a word vector, in
combination with the term frequency, as measure of word significance.

In a discussion forum dedicated to
word2vec,\footnote{http://groups.google.com/forum/\#!forum/word2vec-toolkit}
it has been argued by some that the length of a vector merely reflects
the frequency with which a word appears in the corpus, while others
argued that it in addition reflects the similarity of the contexts in
which a word appears.  According to this thesis, a word that is
consistently used in a similar context will be represented by a longer
vector than a word of the same frequency that is used in different
contexts.  Below, we provide experimental support for this thesis.  It
is this property that justifies measuring significance by word vector
length, for words represented by long vectors refer to a distinctive
context.

It is further proposed that the scatter plot of word vector length
versus term frequency of all the words in the vocabulary provides a
useful two-dimensional visualization of a text corpus.

The paper is organized as follows.  The next section introduces
the language corpus used and gives a global characterization based on
term frequency.  Section \ref{sec:exp} describes the experiments carried
out using word2vec, and presents the main results.  Section
\ref{sec:disc} concludes the paper with a short discussion.  

\section{Dataset}
\label{sec:corpus}
For our experiments we use a dataset from the
arXiv,\footnote{http://arxiv.org/} a repository of scientific preprints.
The dataset consists of about 29k papers from one single subject class
in the arXiv, \textit{viz.}\ the hep-th section on theoretical
high-energy physics posted in the period from January 1992 to April
2003.  Although full papers are available, we consider only title and
abstract of the papers, which have about 100 word tokens on
average.\footnote{The dataset is available from the KDD Cup 2003
  homepage
  http://www.sigkdd.org/kdd-cup-2003-network-mining-and-usage-log-analysis}

LaTeX (or TeX) commands are removed from input through use of the detex
program.\footnote{http://www.cs.purdue.edu/homes/trinkle/detex/} The
input text is further converted to lowercase, and punctuation marks and
special symbols are separated from words, as was done in the
preprocessing step of a word2vec experiment by Mikolov on the IMDB
dataset of movie
reviews.\footnote{https://groups.google.com/forum/\#!msg/word2vec-toolkit/Q49FIrNOQRo/J6KG8mUj45sJ}

\subsection{Term Frequency List}
\begin{table}[t]
\begin{center}
\begin{tabular}{lrr}
term & $v$ & $tf$ \\ \hline
theory & 1.90 & 27702 \\
field & 2.00 & 17510 \\
gauge & 2.13 & 15536 \\
string & 2.33 & 13523 \\
model & 2.19 & 12389 \\
quantum & 2.14 & 12307 \\
theories & 2.22 & 10528 \\
space & 2.18 & 8035 \\
also & 1.67 & 7907 \\
models & 2.39 & 7313 \\
two & 2.03 & 7286 \\
fields & 2.11 & 7261 \\
solutions & 2.49 & 7129 \\
show & 1.87 & 7125 \\
action & 2.39 & 6602 \\
one & 1.82 & 6440 \\
black & 3.24 & 6011 \\
dimensions & 2.34 & 5953 \\ 
symmetry & 2.35 & 5792 \\
group & 2.46 & 5696 \\
equations & 2.54 & 5509 \\
algebra & 2.70 & 5461
\end{tabular}
\caption{Top ranked words in the term frequency list of the hep-th corpus
  with their vector length $v$ (included for later convenience) and term
  frequency $tf$.  Punctuation marks and stop words are removed from the
  list.}
\label{table:top}
\end{center}
\end{table}
After removing stop words and punctuation marks, the list of 50 most
frequently used words in the corpus reduces to the one given in
Table~\ref{table:top}.  Deriving from a domain-specific dataset, this
list indeed gives a succinct and fairly precise characterization of the
hep-th corpus, which is primarily about ``gauge theory'', ``quantum
field theories'', and ``string theory''.  It correctly reveals the
importance of ``models'' in this research area, as well as the
importance of the concepts ``space'', ``solutions'', ``action'',
``dimensions'', ``symmetry group'', ``equations'', and ``algebra''.  The
term ``black'' refers to ``black holes'', which play a distinctive role
in this corpus.  The term ``also'' is not filtered out by the
NLTK\footnote{www.nltk.org/} stop word list we use.  Finally, ``show''
(used exclusively as verb in this corpus, not as noun) appears mostly in
the context ``we show that'' and reveals that a large portion of the
corpus consists of research papers.  Note that ``show'' is the only verb
besides the stop word ``be'' that made it into the top 50 list.

\section{Experiments}
\label{sec:exp}
We next turn to word2vec.\footnote{The code is available for download at
  https:// code.google.com/p/word2vec} For training the neural network,
we use the same parameter settings as advertised for the IMDB dataset
referred to above.\footnote{Specifically, the parameters used are:
  \texttt{word2vec -train \$inputfile -output \$outputfile -cbow 0 -size
    100 -window 10 -negative 5 -hs 0 -sample 1e-4 -threads 40 -binary 0
    -iter 20 -min-count 1}} With these settings, the vector dimension is
100, the (maximum) context window size is 10, and the algorithm makes 20
passes through the dataset for learning.  The total number of tokens
processed by the algorithm is 3.2M.  As is typical for a highly specific
domain, the vocabulary is relatively small, containing about 44k terms,
of which about half is used only once.

\subsection{Similarity Distribution}
During training, similar words are grouped together in the same
direction by the learning algorithm, so that after training the vectors
encode word semantics.  One of the most popular measures of semantic
similarity in NLP is the cosine similarity given by the dot product
between two \textit{normalized} vectors.  Denoting the cosine of the
angle between the two vectors, the cosine similarity can take values in
the interval $[-1,1]$. 

\begin{figure}
\begin{center}
  \includegraphics[width=.4\textwidth]{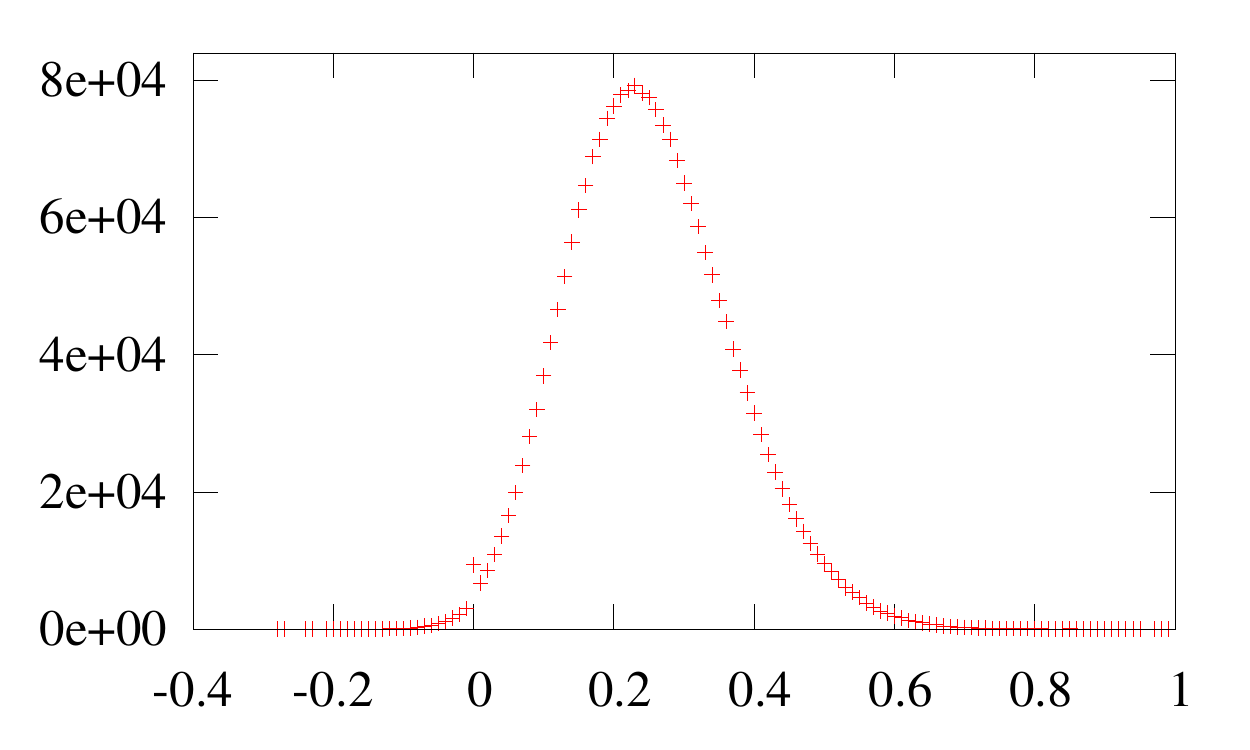}
\end{center}
\caption{Cosine similarity between arbitrarily chosen pairs of word
  vectors with $tf>1$.}
\label{fig:sim}
\end{figure}

To analyze the hep-th corpus, we built a histogram of the cosine
similarity between arbitrarily chosen pairs of word vectors.  The words
are randomly selected from the vocabulary irrespective their frequency.
We have, however, discarded terms that appear only once.  The result,
given in Fig.~\ref{fig:sim}, is a bell-shaped distribution.
To our surprise, the distribution is not centered around zero, but
around a positive value, $0.23$.  This means that word vectors in the
hep-th corpus have on average a certain similarity.  Closely related to
this is that the average word vector is non-zero, having a small length,
$v = 1.37$ ($v=1.51$ when words that only appear once are excluded).
This vector marks the center of the word cloud spanned in the word
vector space by all the words in the vocabulary.

To see if this behavior is shared by general purpose corpora, we
considered Wikipedia by way of example.\footnote{For a cleaned version
  of the Wikipedia corpus from October 2013, see
  https://blog.lateral.io/2015/06/the-unknown-perils-of-mining-wikipedia/}
For that corpus, covering diverse topics, we found, as expected, the
histogram to be centered around zero (and slightly right skewed).  The
non-zero value found for the hep-th corpus is therefore probably a sign
of the homogeneity of this dataset.

The reason for excluding terms that only appear once, which after all
make up half of the vocabulary, is that they have their own bell-shaped
distribution, slightly more peaked than the one shown in
Fig.~\ref{fig:sim} and centered at a higher cosine similarity of about
$0.5$.

Note the outlier at zero cosine similarity in the distribution in
Fig.~\ref{fig:sim}.  The reason for this outlier eludes us.

\subsection{Vector Length as Significance Factor}
\begin{figure}
\begin{center}
  \includegraphics[width=.4\textwidth]{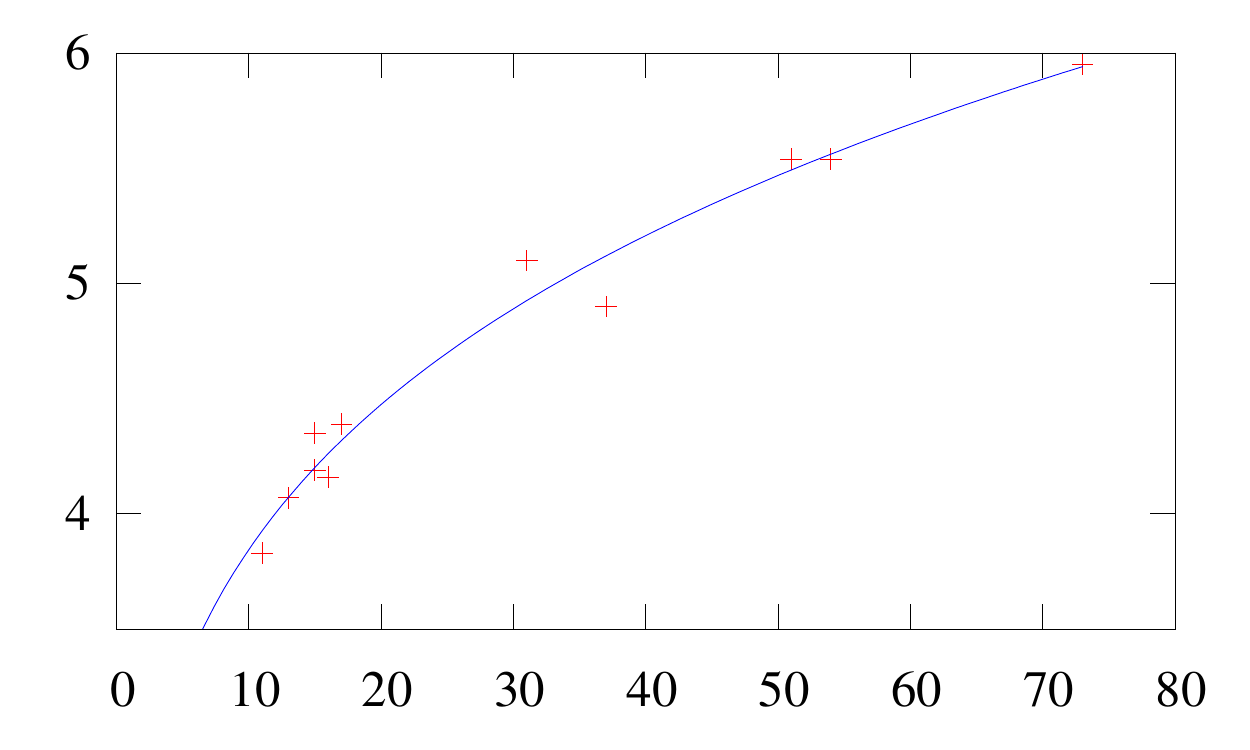}
\end{center}
\caption{Word vector length as a function of frequency of appearance of
  the months of the year excluding ``may''.  The line through the data
  points serves as a guide to the eye.}
\label{fig:months}
\end{figure}

To demonstrate that, besides depending on consistent use, the length of
a word vector also depends on term frequency, we consider the months of
the year, see Table~\ref{table:months}.  Apart from the word token
``may'', which in addition denotes a verbal auxiliary and is therefore
used in many different contexts, these terms consistently appear in the
abstracts of the hep-th corpus to indicate the time of a school or
conference where the paper was presented.  The data clearly show that
for fixed context, the vector length increases with term frequency, see
Fig.~\ref{fig:months}.
\begin{table}
\begin{center}
\begin{tabular}{lrr}
month     &   $v$  & $tf$ \\ \hline
january   &   4.16 & 16 \\
february  &   4.19 & 15 \\
march     &   4.90 & 37 \\
april     &   4.07 & 13 \\
may       &   2.23 & 2229 \\
june      &   5.95 & 73 \\
july      &   5.54 & 54 \\
august    &   5.10  & 31 \\
september &   5.54  & 51 \\ 
october   &   3.83 & 11 \\ 
november  &   4.35 & 15 \\ 
december  &   4.39 & 17 
\end{tabular}
\caption{The months of the year with their word vector length $v$ and term
  frequency $tf$.}
\label{table:months}
\end{center}
\end{table}
The three terms with the largest vector length are besides ``june'',
``school'' ($v=5.97, \, tf=114$) and ``conference'' ($v=5.89, \,
tf=93$).  If we take vector length as a measure of word significance,
this finding surprisingly supports another thesis by Luhn
\cite{Luhn:1958}, which states that:
\begin{quote}
The more often certain words are found in each other’s company within a
sentence, the more significance may be attributed to each of these
words.
\end{quote}
Here, the phrase ``certain words'' refers to words that
\begin{quote}
a writer normally repeats [$\cdots$] as he advances or varies his
arguments and as he elaborates on an aspect of a subject.
\end{quote}

Table~\ref{table:months} also nicely demonstrates that term frequency
alone does not determine the length of a word vector.  The term ``may''
has a much higher frequency than the other terms in the table, yet it is
represented by the shortest vector.  This is because it is used in the
corpus mostly as a verbal auxiliary in opposing contexts.  When a word
appears in different contexts, its vector gets moved in different
directions during updates.  The final vector then represents some sort
of weighted average over the various contexts.  Averaging over vectors
that point in different directions typically results in a vector that
gets shorter with increasing number of different contexts in which the
word appears.  For words to be used in many different contexts, they
must carry little meaning.  Prime examples of such insignificant words
are high-frequency stop words, which are indeed represented by short
vectors despite their high term frequencies, see Table~\ref{table:stop}.
\begin{table}
\begin{center}
\begin{tabular}{lrr}
term     &   $v$  & $tf$ \\ \hline
the & 1.49 & 257866 \\
of & 1.51 & 148549  \\
. & 1.43 & 131022   \\
, & 1.41 & 84595    \\
in & 1.59 & 80056   \\
a & 1.61 & 72959    \\
and & 1.51 & 71170  \\
to & 1.61 & 53265   \\
we & 1.88 & 49756   \\
is & 1.69 & 49446   \\ 
for & 1.62 & 34970  \\
) & 2.03 & 28878
\end{tabular}
\caption{Top 12 terms in the term frequency list of the hep-th
  corpus with their word vector length $v$ and term frequency $tf$.  In
  addition to punctuation marks, this list exclusively features stop
  words.}
\label{table:stop}
\end{center}
\end{table}

\subsection{Vector Length vs. Term Frequency}
\begin{figure}[t]
\begin{center}
  \includegraphics[width=.4\textwidth]{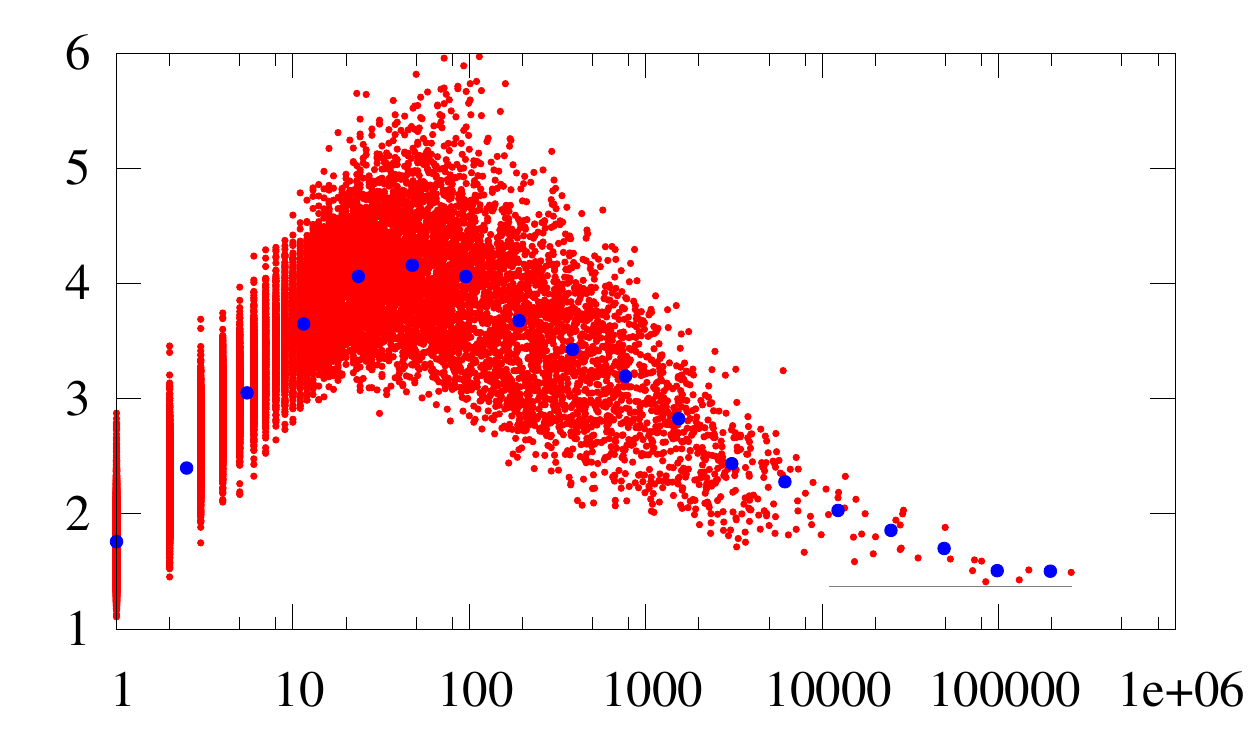}
\end{center}
\caption{Word vector length $v$ versus term frequency $tf$ of all words
  in the hep-th vocabulary.  Note the logarithmic scale used on the
  frequency axis.  The dark symbols denote bin means with the $k$th bin
  containing the frequencies in the interval $[2^{k-1}, 2^k-1]$ with
  $k=1,2,3,\ldots$.  These means are included as a guide to the eye.
  The horizontal line indicates the length $v = 1.37$ of the mean
  vector.}
\label{fig:norm_vs_freq}
\end{figure}

To study to what extent term frequency and word vector length can serve
as indicators of a word's significance, we represent all words in the
vocabulary in a two-dimensional scatter plot using these variables as
coordinates.  Figure~\ref{fig:norm_vs_freq} gives the result for the
hep-th corpus.
For given term frequency, the vector length is seen to take values only
in a narrow interval.  That interval initially shifts upwards with
increasing frequency.  Around a frequency of about 30, that trend
reverses and the interval shifts downwards.  

Both forces determining the length of a word vector are seen at work
here.  Small-frequency words tend to be used consistently, so that the
more frequently such words appear, the longer their vectors.  This
tendency is reflected by the upwards trend in
Fig.~\ref{fig:norm_vs_freq} at low frequencies.  High-frequency words,
on the other hand, tend to be used in many different contexts, the more
so, the more frequently they occur.  The averaging over an increasing
number of different contexts shortens the vectors representing such
words.  This tendency is clearly reflected by the downwards trend in
Fig.~\ref{fig:norm_vs_freq} at high frequencies, culminating in
punctuation marks and stop words with short vectors at the very end.

Words represented by the longest vectors in a given frequency bin often
carry the content of distinctive contexts.  Typically, these contexts
are topic-wise not at the core of the corpus but more on the outskirts.
For example, the words with the longest vector in the high-frequency
ranges $[2^{k-1}, 2^k-1]$ with $k=9,10,11$, ``inflation'' ($v=4.64, \,
tf=571$), ``sitter'' ($v=3.81, \, tf=1490$) as in ``de Sitter'', and
``holes'' ($v=3.41, \, tf=2465$) as in ``black holes'', all refer to
general relativity.  General relativity, having its own subject class
(gr-qc) in the arXiv, is not one of the main subjects of the hep-th
corpus.  It takes a distinctive position in this corpus as it mostly
appears in studies that aim at reconciling general relativity with the
laws of quantum mechanics.
\subsection{POS Tagging}
To further assess the ability of word vector length to measure word
significance, we assign part-of-speech (POS) tags to each word in the
corpus.  For this task, we use the Stanford POS
tagger\footnote{Specifically, we use the
  english-caseless-left3words-distsim tagger.  For details on this model
  and for downloading, see
  http://nlp.stanford.edu/software/tagger.shtml.} \cite{Toutanova:2003}.
The final tag assigned to a word in the vocabulary is decided by
majority vote.  

\begin{figure}
\begin{center}
  \includegraphics[width=.4\textwidth]{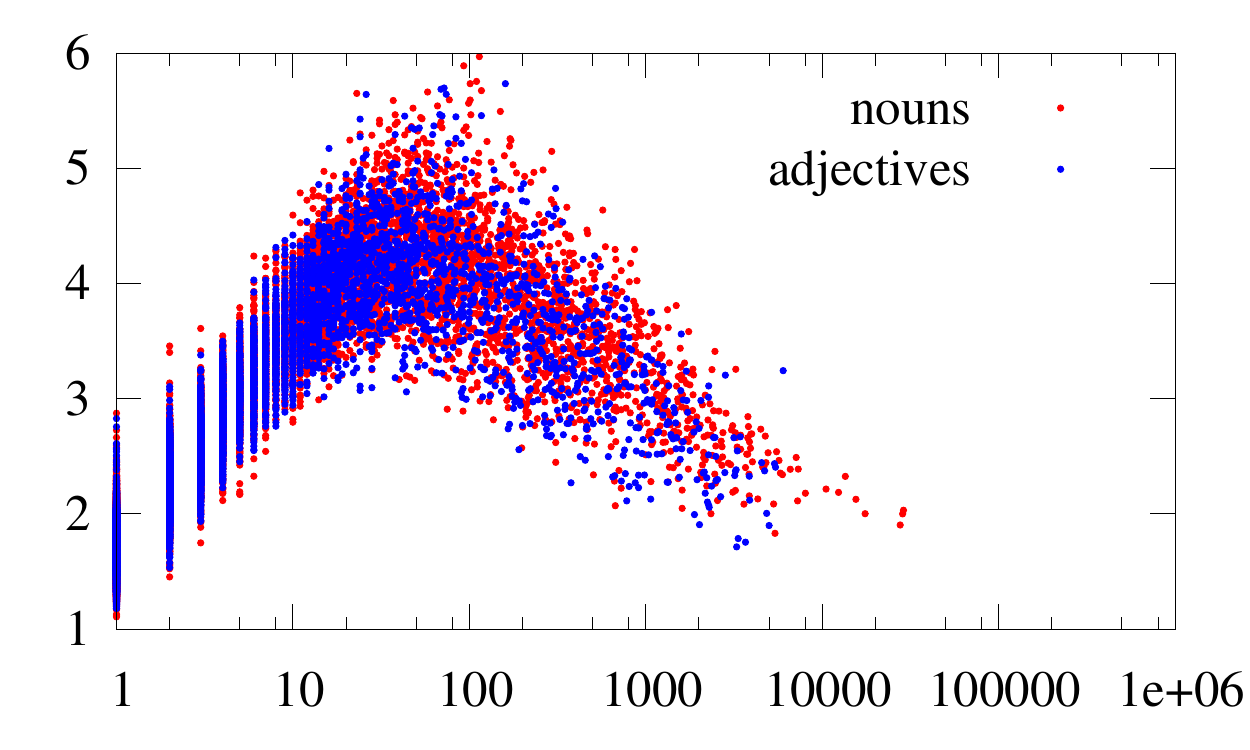}
\end{center}
\caption{Word vector length $v$ versus term frequency $tf$ of all words
  in the hep-th vocabulary labeled nouns (red dots) or adjectives (blue
  dots).}
\label{fig:nn_vs_jj}
\end{figure}

By the way that word2vec learns word representations, we expect nouns
(excluding proper nouns) and adjectives to be similarly distributed in
the $v$-$tf$ plane.  This is indeed what we observe, see
Fig.~\ref{fig:nn_vs_jj}.  Note that these word types pervade almost the
entire region covered in the full plot in Fig.~\ref{fig:norm_vs_freq} by
the complete vocabulary.

\begin{figure}[t]
\begin{center}
  \includegraphics[width=.4\textwidth]{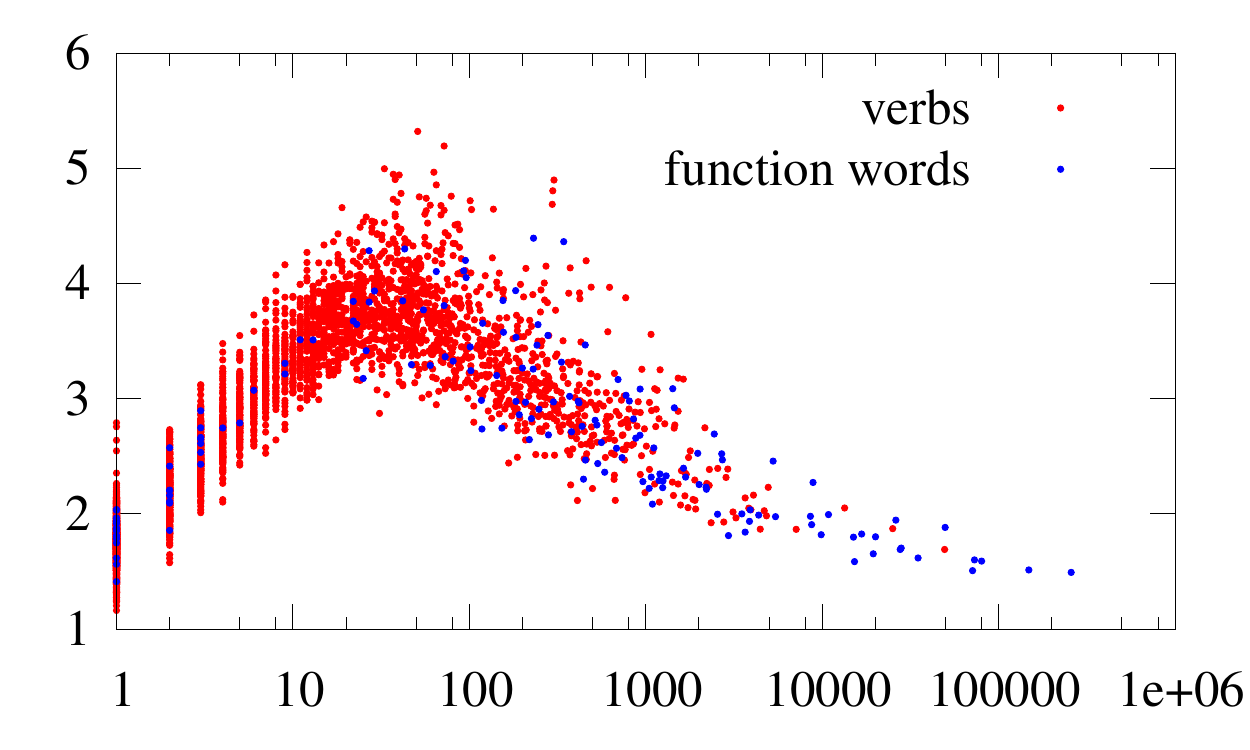}
\end{center}
\caption{Word vector length $v$ versus term frequency $tf$ of all words
  in the hep-th vocabulary labeled verbs (red dots) or function words
  (blue dots).}
\label{fig:vb_vs_fw}
\end{figure}

We also find verbs and adverbs to be similarly distributed in the
$v$-$tf$ plane.  Again this was to be expected given that word2vec
learns word representations from word co-occurrences.  Somewhat
surprisingly, we observe in Fig.~\ref{fig:vb_vs_fw} that the
distribution of verbs also overlaps with that of function
words.\footnote{As function words we classify: prepositions (IN),
  pronouns (PRP, PRP\$, WP, WP\$), determiners (DT, PDT, WDT),
  conjunctions (CC), modal verbs (MD), and particles (RP). In brackets,
  we included here the tags used by the Stanford POS tagger.} These word
types no longer pervade the entire region covered in the full plot, but
are confined to the bottom band, corresponding to short vectors.  The
fact that function words are represented by short vectors underscores
the ability of vector length to measure word significance.

\begin{figure}[t]
\begin{center}
  \includegraphics[width=.4\textwidth]{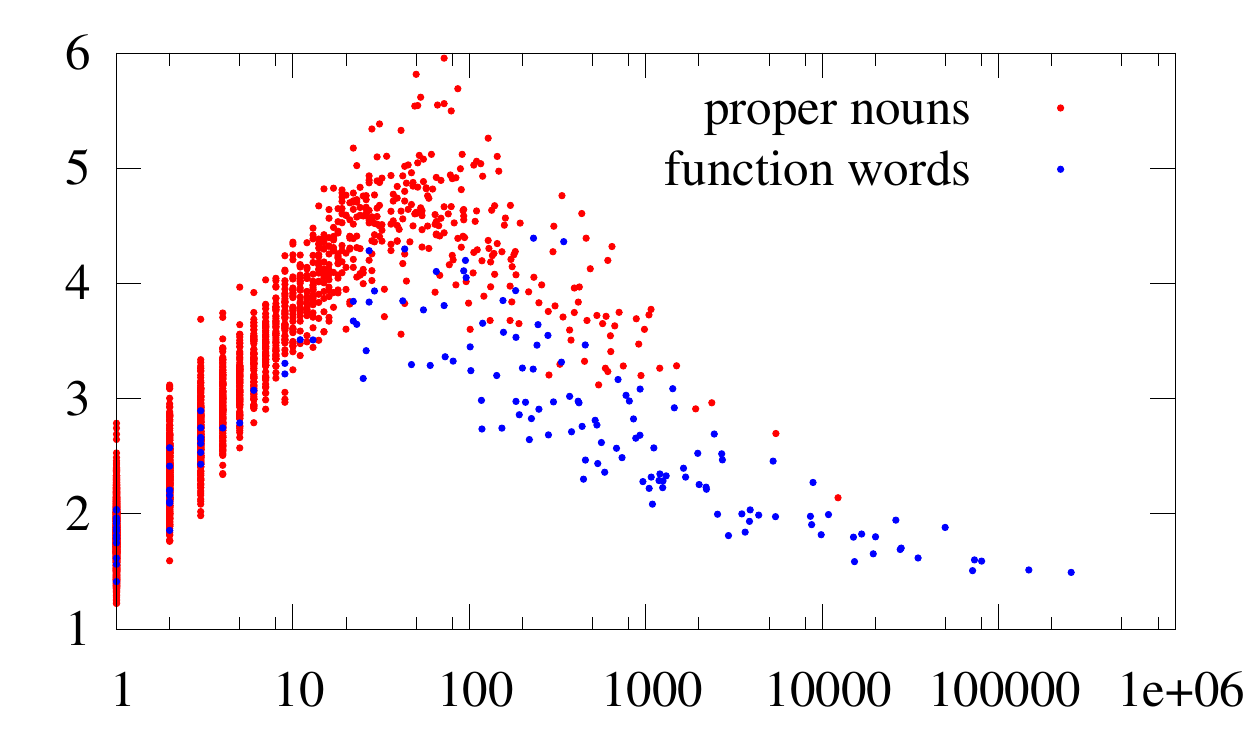}
\end{center}
\caption{Word vector length $v$ versus term frequency $tf$ of all words
  in the hep-th vocabulary labeled proper nouns (red dots) or function
  words (blue dots).}
\label{fig:nnp_vs_fw}
\end{figure}

This proficiency is even more brought out by comparing the distribution
of function words and proper nouns, which typically are indicative of
distinctive contexts in the hep-th corpus.  The plot in
Fig.~\ref{fig:nnp_vs_fw} shows a clear separation of proper nouns and
function words for sufficiently large term frequencies.

\subsection{Visualization}
These results suggest an interesting technique for visualizing text
corpora.  By labeling the data points in the $v$-$tf$ scatter plot with
their terms, one obtains a two-dimensional visualization of all the
words in the vocabulary.  One advantage of a $v$-$tf$ plot is that words
are ranked by significance, thus allowing for effective exploration of a
corpus.  To deal with the large number of data points, an interactive
visualization tool can be build that allows the user to navigate a mouse
pointer over the plot, and that shows only the labels of the data points
in the vicinity of the pointer.

There exist other techniques for visualizing high-dimensional data, such
as the popular t-distributed stochastic neighbor embedding (t-SNE)
\cite{vdmaaten:2008}.  That machine learning algorithm, being an example
of multidimensional scaling, projects high-dimensional data points onto
a plane such that the distances, or similarities between them are
preserved as well as possible.  Words of similar meaning thus tend to be
projected together by the t-SNE algorithm.  Since the cosine similarity
is independent of vector lengths, word significance is ignored when
using this measure.  The t-SNE algorithm therefore arranges the data
points entirely differently from our proposal.  Moreover, in contrast to
the axes in the $v$-$tf$ scatter plot, those in the t-SNE plot have no
direct meaning.

\section{Discussion}
\label{sec:disc}
Most applications of distributed representations of words obtained
through word2vec so far centered around semantics.  A host of
experiments have demonstrated the extent to which the direction of word
vectors captures semantics.  In this brief report, it was pointed out
that not only the direction, but also the length of word vectors
carries important information.  Specifically, it was shown that word
vector length furnishes, in combination with term frequency, a useful
measure of word significance.  Also an alternative to the t-SNE
algorithm for projecting word vectors onto a plane was introduced, where
words are ordered by significance rather than similarity.

We have restricted ourselves to unigrams in this exploratory study.  For
more extended experiments and applications, including important bi- and
tri-grams into the vocabulary will certainly improve results.  We have
also restricted ourselves to running word2vec using parameters that were
recommended by the developers, and have not attempted to optimize them.

Finally, the question arises whether word vectors produced by other
highly scalable machine learning algorithms built on top of word
co-occurrences, such as the log bilinear model \cite{Mnih:2013} and GloVe
\cite{pennington:2014}, also encode word significance in their length.

\bibliographystyle{acl} 
\bibliography{unsup_NLP}
\end{document}